\documentclass[10pt,twocolumn,letterpaper]{article}

\usepackage{cvm}
\usepackage{times}
\usepackage{epsfig}
\usepackage{graphicx}
\usepackage{booktabs}
\usepackage{amsmath} 
\usepackage{amssymb}
\usepackage[ruled]{algorithm2e} 
\usepackage{multirow}
\usepackage{multicol}
\usepackage{graphicx}
\usepackage{subfig}
\usepackage{comment}
\usepackage{animate}

\usepackage[pagebackref=true,breaklinks=true,letterpaper=true,colorlinks,bookmarks=false]{hyperref}

\newcommand{\keywords}[1]{{\bf \emph{Keywords: #1}}}

\cvmfinalcopy

\def\cvmPaperID{****} 

\ifcvmfinal\fi
\begin{document}

\title{WonderVerse: Video Geometric Correction and Enhancement for Extendable 3D Scene Generation}

\author{Hao Feng$^{1}$\thanks{Equal contribution.} \quad 
Zhi Zuo$^{1}$\footnotemark[1] \quad  
Jia-Hui Pan$^{2}$\quad 
Ka-Hei Hui$^{3}$ \quad
Qi Dou$^{2}$\quad 
Jingyu Hu$^{2}$\thanks{Corresponding authors.}\quad 
Zhengzhe Liu$^{1}$\footnotemark[2]\\
$^1$Lingnan University\quad 
$^2$The Chinese University of Hong Kong \quad 
$^3$AutoDesk Research \quad 
}
\maketitle

\begin{figure*}
    \centering
    \includegraphics[width=1\linewidth, height=0.42\linewidth ]{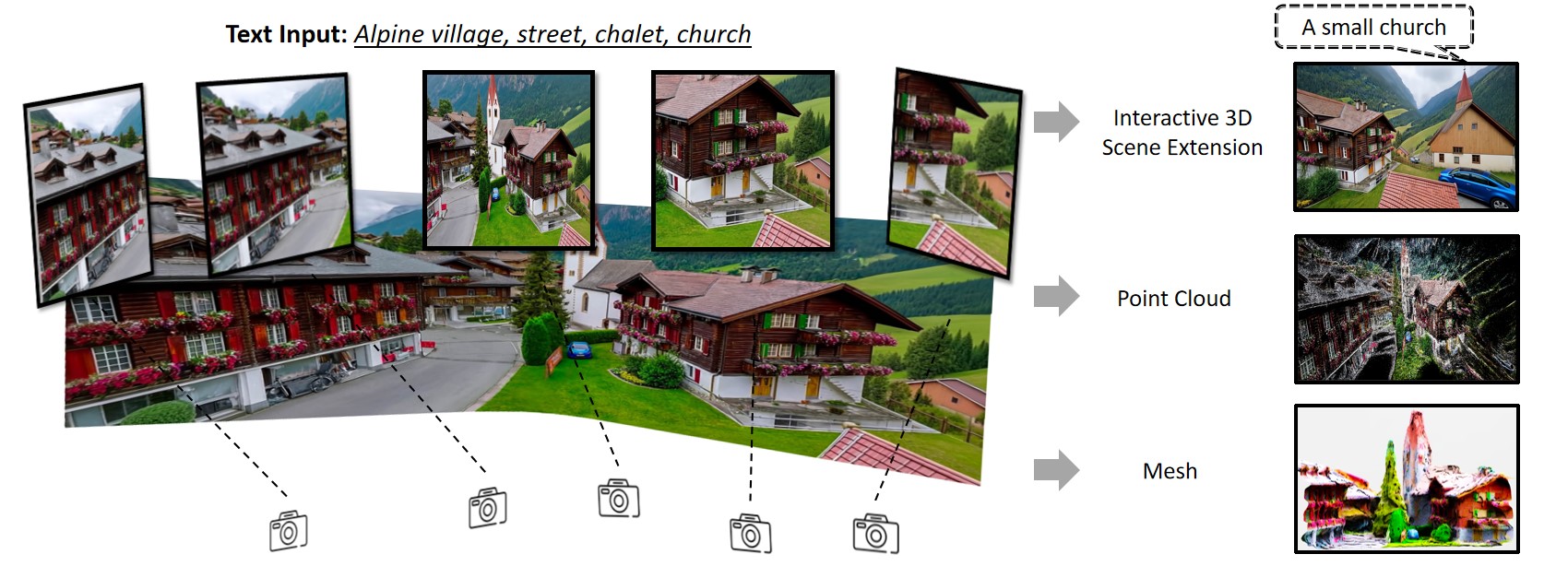}
     \caption{WonderVerse can create large-scale, coherent, extendable, and high-quality 3D scenes from a text. It also allows users to interactively extend the 3D scene with text and can output the generated 3D scene with different representations, including a point cloud and a mesh.}  
     \label{fig:teaser}
\end{figure*}

\begin{abstract}

We introduce \textit{WonderVerse}, a simple but effective framework for generating extendable 3D scenes.  Unlike existing methods that rely on iterative depth estimation and image inpainting, often leading to geometric distortions and inconsistencies, WonderVerse leverages video generative foundation models to create highly immersive, geometrically coherent, extendable, and large-scale 3D environments. 
For enhanced scene completeness and user-driven expansion, an LLM-driven extension module enables semantic expansion and personalization of 3D scenes based on user intent. 
To bridge the video-3D domain gap and enhance video geometric fidelity, WonderVerse incorporates two key modules: a video geometric correction module, which filters geometrically inconsistent abnormal videos by analyzing camera trajectory smoothness, and a 3D geometric enhancement module, which enforces depth and normal consistency between the generated video and the reconstructed 3D scene. 
Extensive experiments on 3D scene generation demonstrate that our WonderVerse, with an elegant and simple pipeline, delivers extendable and highly realistic 3D scenes, markedly outperforming existing works that rely on more complex architectures. The code is available at \href{https://github.com/FH2070/WonderVerse}{https://github.com/FH2070/WonderVerse}
\end{abstract}


\section{Introduction}
\label{sec:intro}

3D scene generation is a fundamental task in the vision community due to its wide-ranging applications in Virtual Reality, Mixed Reality (VR/MR), robotics, self-driving vehicles, and more.  However, creating extendable and large-scale 3D scenes poses significant challenges. First, there is a lack of large-scale, high-quality datasets of 3D scenes. Second, generating realistic and geometrically accurate 3D scenes is inherently complex due to the unstructured nature of 3D space and the need to capture intricate object relationships and environmental context. Third, seamlessly extending 3D scenes while maintaining coherence remains a significant challenge.

%

To generate extendable 3D scenes, recent studies~\cite{hollein2023text2room,chung2023luciddreamerdomainfreegeneration3d,yu2024wonderjourney,yu2024wonderworld} leverage 2D image generative models for 3D scene generation. These methods typically iteratively extend a given image to synthesize views from new camera poses.  Typically, they first estimate a depth map from the input image, project it into 3D, and then use an image inpainting model to extend the scene from novel viewpoints. This iterative process builds and extends the 3D environment. However, the quality of scenes generated by these approaches remains limited. As shown in Figure~\ref{fig:comparedgeneratedscenes}, a primary issue is geometric distortion, arising from the scale ambiguity of single-view depth estimation and the lack of consistent camera parameters across iterations.  Additionally, error accumulation is a concern due to the iterative application of depth estimation and inpainting during scene expansion.  Furthermore, seams and discontinuities can appear between regions generated in different iterations.  Overall, the iterative discrete-view-based generation pipeline inherently results in unsatisfactory scene quality.

Going beyond existing works that rely on image generation techniques, we propose a simple yet effective approach to leverage the recent advancements of LLM and video generative models~\cite{kong2024hunyuanvideo,chen2025goku,vondrick2016generating,zeng2023makepixelsdancehighdynamic,zheng2024open,yang2024cogvideox} to produce highly immersive and superior-quality 3D scenes without dedicated 3D datasets.
{Importantly, our contribution lies not in proposing new foundation models or reconstruction algorithms, but in demonstrating how recent video generative priors can be reliably harnessed for \emph{extendable 3D scene generation}. Compared to image-based iterative pipelines, this design mitigates geometric distortion, error accumulation, and seam artifacts during iterative expansion.}
To achieve this, a natural initial thought is to combine video generative models with 3DGS (3D Gaussian Splatting)~\cite{kerbl20233d} for 3D scene generation.  Nevertheless, this simple combination does not yield satisfactory results 
since the video generative model cannot faithfully capture the 3D geometric rules of the real world. First, current video generative models often lack the precise controllability required for directional extension in 3D space. 
Second, as indicated by existing works~\cite{qin2024worldsimbench,chang2024matters}, AI-generated videos can be geometrically inconsistent across frames; thus, the contradictions among the frames can lead to the distortion of the created 3D scenes. 
Moreover, the inherent dimensionality reduction of projecting a 3D scene onto a 2D video inevitably leads to a loss of crucial 3D surface detail. This is because subtle geometric variations, which are essential for accurate 3D surface reconstruction, are often compressed or obscured during the video capture process, ultimately resulting in surface degradation in generated videos and inferior 3D surface quality. 

To address these issues, we present WonderVerse, a simple but effective extendable 3D scene generative model by enhancing the geometric fidelity of video generative models. 
First, to make the generated scenes more complete, vivid, and adaptable to diverse user needs, we introduce an LLM-driven extension module. This component leverages large language models to semantically interpret and expand the generated 3D scenes, enabling the generation of richer contextual elements beyond the originally captured content, empowers users to flexibly customize or extend scenes based on high-level descriptions or creative intent, significantly enhancing interactivity and scalability of the generation pipeline.
Second, we present an abnormal sequence detection module to filter out geometrically inconsistent generated videos. 
This design stems from our observation that COLMAP camera pose estimation effectively flags the geometric issue: inconsistent videos produce erratic camera poses and discontinuous trajectories. The module evaluates the continuity of the estimated camera trajectory to enhance video geometric coherence. 
Third, we introduce a 3D geometric enhancement module to provide explicit depth and surface normal supervision during training. By enriching the generated videos with depth and normal maps, we encourage the resulting 3D scenes to align with the videos, enforcing consistency in both RGB values and depth/normal constraints.
With consistent and geometrically enriched video sequences, we reconstruct the 3D scene. Note that our approach is compatible with various 3D reconstruction approaches for both efficient (DUSt3R~\cite{dust3r_cvpr24}) and high-quality (3DGS~\cite{kerbl20233d}) 3D scene generation. 

 As illustrated in Figure~\ref{fig:teaser}, our WonderVerse can create an extendable and highly immersive 3D scene from a piece of text. Furthermore, both qualitative and quantitative experimental results demonstrate that WonderVerse’s neat and simple pipeline leads to state-of-the-art, extendable, and highly realistic 3D scene generation. Our approach demonstrates that sophisticated results can be attained through our highly elegant design.
 \textbf{Codes will be released upon publication. }
%


In summary, the key contributions of WonderVerse are:
\begin{itemize}

    \item We propose a simple yet effective approach to leverage video generative models for the extendable 3D scene generation task, surpassing image-based iterative pipelines.

    \item We design a LLM-driven extension module to extend scenes based on high-level descriptions, significantly enhancing interactivity and scalability of the generation pipeline.
    
    \item We present two novel modules to enhance the geometric fidelity of AI-generated videos: a geometric correction module with an abnormal sequence detection module to ensure geometric consistency and a 3D geometric enhancement module to improve the 3D surface quality.

    \item WonderVerse achieves state-of-the-art extendable and highly-realistic 3D scene generation through a highly neat and simple pipeline, significantly outperforming existing methods with far more complex architectures. 


\end{itemize} 

\section{Related Works}
\label{related works}
In this section, we will briefly introduce the development of  video generation for 3D reconstruction, extendable 3D scene generation, and normal-depth estimation. And we also introduce novel view synthesis (NVS) in supplementary material.

\subsection{Image and Video Generative Models for MVS and 3D Reconstruction}

%
%
%
Recent progress in foundation models for image~\cite{rombach2022high} and video~\cite{blattmann2023stable} generation has driven the exploration of their use in 3D reconstruction. Image generative models can be fine-tuned with multi-view data for novel-view synthesis and 3D reconstruction of objects~\cite{liu2023zero,long2024wonder3d} and scenes~\cite{gao2024cat3d,wu2024reconfusion}. This strategy has been extended to video foundation models, enabling video generation with controlled camera motion for 3D reconstruction~\cite{bar2024lumiere,vondrick2016generating,li2024dreamscene,voleti2024sv3d,xie2024sv4d,he2024cameractrl,wang2024motionctrl}. Besides, recent advancements in 3D generative models~\cite{liu2023exim, hui2022neural, hu2023neural}, have demonstrated impressive capabilities in controllable 3D generation~\cite{hu2023clipxplore, hu2024_cnsedit, hu2026pegasus3dpersonalizationgeometry, yan2026comp, hui2022template} and analysis~\cite{du2025hierarchical}.
Nevertheless, these methods primarily target object-level or limited-scale reconstruction and cannot directly adapt to our extendable 3D scene generation task. 
Our objective is to generate extendable 3D scenes, requiring models with the imaginative ability to extend 3D scenes beyond the initial input.

\vspace{-2mm}

\subsection{Extendable 3D Scene Generation}

Another branch of works explore extendable 3D scene generation. 
Existing works like~\cite{xie2024citydreamer, xie2024gaussiancity,zhang2024cityx,deng2023citygen,lin2023infinicity} only focus on urban environments and cannot be generalized to other scenes, restricting their semantic scope. For unbounded nature scenes, methods like~\cite{chen2023scenedreamer,chai2023persistent} require multi-view data, while others~\cite{wu2024blockfusion,liu2024pyramid,wei2024planner3d} rely on 3D scene datasets. To avoid such data dependencies, Text2Room~\cite{hollein2023text2room}, SceneScape~\cite{fridman2023scenescape}, VividDream~\cite{lee2024vividdream} LucidDreamer~\cite{chung2023luciddreamerdomainfreegeneration3d}, WonderJourney~\cite{yu2024wonderjourney}, WonderWorld~\cite{yu2024wonderworld} generate scenes iteratively with depth estimation and inpainting. Among existing approaches, \textbf{these methods are the most relevant ones to our work. } However, their iterative pipelines inherently suffer from geometric distortion, discontinuity, and error accumulation during scene expansion, ultimately hampering scene quality and motivating our WonderVerse framework.
In this work, we seek to overcome the above challenges and leverage video generative models to generate more immersive and expandable scenes.

\vspace{-2mm}

\subsection{Normal and Depth Priors for 3D Generation}
%
%
Recent work in monocular depth estimation~\cite{yang2024depth, yang2024depth_v2, yin2023metric3d, hu2024metric3d, ke2024repurposing, chen2025video, hu2024depthcrafter} has shown that pre-trained models can reliably predict depth priors from images or videos.
%
%
Recently, the integration of geometric priors, i.e., normal and depth, for 3D generation has emerged as an important strategy to alleviate the inherent ambiguities in reconstructing 3D structures.
Existing methods~\cite{ye2024stablenormal, bae2024rethinking} focus on extracting surface normals directly from images, providing complementary geometric details that improve downstream tasks such as 3D reconstruction and scene understanding.
In particular,~\cite{qiu2024richdreamer, fu2024geowizard} aim to unify these approaches by jointly predicting depth and surface normals.
Moreover, recent methods~\cite{long2024wonder3d, ye2025hi3dgen} explicitly leverage depth and normal maps as supervision signals during training, improving the fidelity of generated 3D assets.
%
While methods like~\cite{long2024wonder3d, ye2025hi3dgen} achieve impressive results in object-level 3D generation, they rely on ground truth depth and normal maps, which are expensive to obtain for 3D scenes. This problem hinders their direct extension to scene generation.
To address this limitation, our approach leverages pre-trained models~\cite{yang2024depth_v2, ye2024stablenormal} to extract robust depth and normal priors, which serve as foundational guidance for 3D scene generation while bypassing the need for ground-truth annotations. 

\begin{figure*}
    \centering
    \includegraphics[width=0.99\linewidth, height=0.45\linewidth]{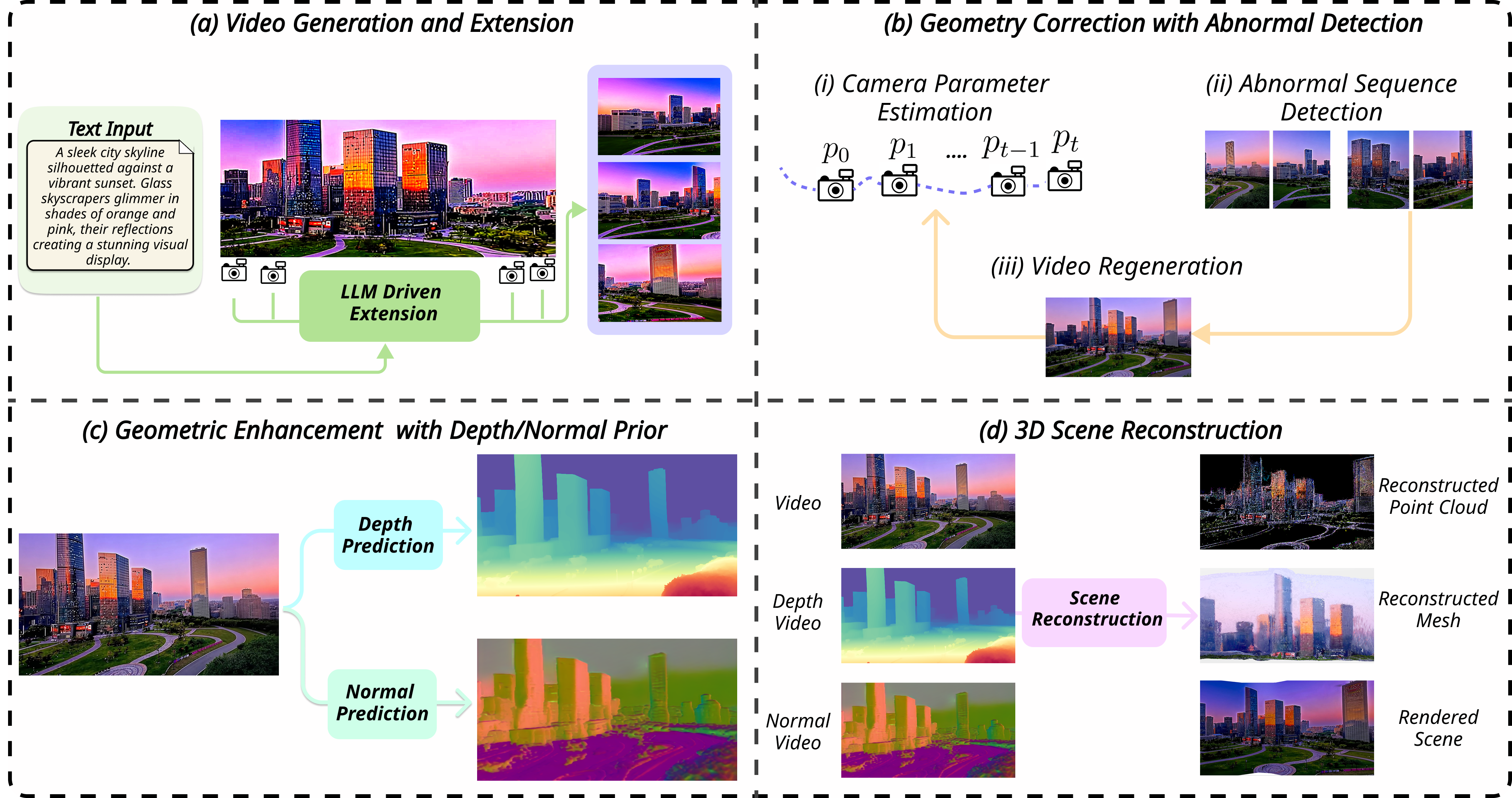}
     \caption{\textbf{Illustration of our WonderVerse.} 
     %
     This framework includes: (a) a text-guided video generation and LLM driven extension module that produces a video of a scene circularly in a continuous shot, followed by extensions in any direction; (b) geometry correction module with abnormal detection, designed to maintain cross-frame 3D geometric coherence in video sequences; (c) a geometric enhancement module that integrates pre-trained depth/normal prediction models to generate depth/normal maps, establishing robust geometric priors and improving the subsequent 3D scene reconstruction; and (d) a 3D scene reconstruction module that processes generated video sequences alongside predicted depth/normal maps as input, simultaneously producing both photorealistic renderings and geometrically consistent 3D reconstructions.
     }
     \label{fig:framework}
\end{figure*}

\vspace{-2mm}
\section{Methods}
In this section, we will introduce our \textit{WonderVerse} which is illustrated in Figure~\ref{fig:framework}.
Given a text description as input, WonderVerse first generates a video through the text-guided video generation and extension module (Figure~\ref{fig:framework} (a)), which continuously presents the scene and extends it to the left and right sides. To ensure the geometric consistency of the video sequence, the Geometric Correction module (Figure~\ref{fig:framework} (b)) estimates the camera pose sequence, detects abnormal sequences by identifying pose discontinuities, and regenerates them. Simultaneously, the Geometric Enhancement module (Figure~\ref{fig:framework} (c)) predicts depth maps and normal maps for the video frames to provide important geometric prior information. Finally, the 3D Scene Reconstruction module (Figure~\ref{fig:framework} (d)) utilizes the geometrically corrected and enhanced video sequences for 3D scene reconstruction, thereby building the complete scene. 

\vspace{-3mm}
\subsection{Video Generation and LLM Driven Extension} 

Most existing works on extendable 3D scene generation rely on iterative image generation.
Such a design can hinder the quality of the generated results due to the significant domain gap between images and 3D representations.
%
%
Inspired by the rapid development of video generative models like~\cite{kong2024hunyuanvideo}, we reconsider the problem by leveraging the superior generative capability of such models.  
Therefore, we propose to address the extendable 3D scene generation task through the lens of video generation.

Given a text prompt $P$, 
we use text-guided video generation model~\cite{kong2024hunyuanvideo} to obtain an initial video $V_\text{init}\in \mathbb{R}^{T\times H\times W \times 3}$, where $H$ and $W$ denote the height and width of each frame, respectively. 
$T$ denotes the length of the video and
$3$ denotes RGB channels. 
%
%
To extend the scene, we further perform video-based scene extensions by extending the initial video.
%
%
To maintain style consistency, we use the first frame $V_\text{init}^{t=1}$ and the last frame $V_\text{init}^{t=T}$ of the initial video as references and generate new videos to extend the scene in any direction. Here we use the left and right extensions as an example. We generate two new videos $V_\text{extend}^\text{left} \in \mathbb{R}^{T'\times H\times W \times 3}$ and $V_\text{extend}^\text{right} \in \mathbb{R}^{T'\times H\times W \times 3}$. 
Then, we combine them into an extended video $\{V_\text{extend}^\text{left}, V_\text{init}, V_\text{extend}^\text{right}\}$ with length $T+2T'$.
The video extension can be applied repeatedly $n$ times, taking $V_\text{input}$ as the initial video for the next iteration, further enriching the scene and yielding $V_\text{input}= \{V_\text{extend}^{\text{left}_{n}}, ..., V_\text{init}, ..., V_\text{extend}^{\text{right}_{n-1}}, V_\text{extend}^{\text{right}_{n}}\} = \{V_{1},V_{2},...,V_{2n+1}\}$.
Figure~\ref{fig:framework} (a) illustrates two iterations of video extension. To achieve reasonable and natural scene expansion, we propose an LLM-driven extension module that generates relative camera trajectories based on user-provided prompts. Specifically, given a textual scene description, the module predicts plausible camera motions to explore newly imagined or extended areas of the scene. Furthermore, our system supports iterative language-based interaction, allowing users to progressively expand the scene by issuing high-level natural language commands. In each iteration, the LLM generates a new camera trajectory segment conditioned on both the current scene context and the user’s intent, enabling flexible and controllable scene expansion. 
More details can be found in Experiments Section~\ref{experiments}.
%

\vspace{-2mm}
{Despite the high quality of AI-generated videos, geometric inconsistencies across frames can still occur, which significantly degrades downstream 3D reconstruction.}

{We empirically observe a strong correlation between visually evident geometric failures and discontinuous camera trajectories estimated by Structure-from-Motion (SfM).}
{Motivated by this observation, we introduce an abnormal sequence detection module that acts as a lightweight \emph{sampling/filtration} mechanism (rather than a theoretical guarantee of geometric correctness) to discard clearly degenerate video segments before reconstruction.}

{First, we perform \textit{camera pose estimation} for each frame in the generated videos $\{V_{1}, V_{2},..., V_{2n+1}\}$ using COLMAP (SfM), producing an estimated camera trajectory per video segment.}

{Second, we \textit{detect abnormal sequences} by checking trajectory discontinuities: a frame is marked as discontinuous if its translation jump exceeds $T_1$ or its rotation change exceeds $T_2$ (relative to the previous frame). If any discontinuity is detected, the corresponding video segment is treated as abnormal.}

{Finally, we perform \textit{geometric correction} via \emph{full-segment regeneration}: once an abnormal segment is detected, we regenerate the \emph{entire} video segment (instead of locally repairing a few frames or partially patching trajectories). Each regeneration is sampled independently with a new random seed, which avoids local patching artifacts and prevents error accumulation across retries.}

{In practice, this simple strategy is effective and does not lead to excessive retry loops: as reported in Sec.~3.2, the average number of additional trials is only 0.25.}
{We also note a limitation: SfM-based pose estimation can miss subtle geometric warping or occasionally produce false positives; therefore, our module is not intended to ``fix'' the video generator, but to reliably filter out failure cases that would inevitably break the 3D reconstruction stage.}
An algorithm outlining the process is provided in the supplemental material.

\vspace{-3mm}
\subsection{Geometric Enhancement with Depth and Normal Priors}
While reducing the 2D-to-3D domain gap compared to static images, videos still lack essential 3D geometric information like depth and surface normals. 
%
To address this issue, we introduce a video geometric enrichment module that predicts depth and normals from video frames to guide 3D reconstruction training.
We utilize Depth Anything V2~\cite{yang2024depth_v2} for accurate monocular depth estimation, since it is effective even in challenging scenes and offers strong geometric cues.
Besides, to improve 3D detail and surface refinement, we apply the Dense Prediction Transformer (DPT)~\cite{ranftl2021vision} with a hybrid backbone to predict per-frame surface normals, yielding smooth and consistent normals. 
These predicted depths and normals serve as explicit geometric supervision, encouraging geometrically consistent 3D scene reconstruction and enhancing surface quality and detail.

\vspace{-3mm}

\subsection{Geometric-Aware 3D Scene Reconstruction}

With the geometrically validated and enhanced video sequences and their camera poses, we then optimize a 3DGS~\cite{kerbl20233d} representation of the scene. Please refer to the supplementary material for preliminaries of 3DGS. 


\vspace{-2mm}
\paragraph{Depth from 3DGS}
To incorporate the geometric-aware optimization, we estimate per-pixel depth rendered using the discrete volume rendering approximation, similar to color rendering of 3DGS: 
\begin{equation}
\hat{D} = \sum_{i \in N} {{d}}_{i}\alpha_i\prod_{j = 1}^{i - 1} (1- \alpha_j),
\label{eq:depth}
\end{equation}
where $d_i$ is the $i^{th}$ Gaussian depth coordinate in view space and $\alpha$ is opacity. 
To resolve scale ambiguity in monocular depth estimation, we align its scale with that of the 3D Gaussian Splatting (3DGS) depth, as computed by Equation~\ref{eq:depth}. This alignment is achieved by solving for a per-image scale $a$ and shift $b$ using linear regression: 
\begin{equation}
\label {eq:mono-sparse-alignment} \hat {a}, \hat {b} = argmin_{a,b} \sum_{ij} \| (a\hat{D}_{ij} + b) - D_{ij}\|_2^2,
\end{equation}
where $\hat{D}_{ij}$ and $D_{ij}$ denote corresponding per-pixel depth values from 3DGS and monocular depth estimation. 

\paragraph{Surface Normal from 3DGS}
For surface normal estimation, 
the minimal axis of each Gaussian serves as an approximation of the normal direction, which can be formulated as:
\begin{equation}
{\boldsymbol {\hat {n}}_i} =   \text{normalize}(argmin (s_{i1}, s_{i2}, s_{i3})),
\end{equation}
where $s_{ij}$ indicates the scale in the $j$ direction of Gaussian $i$.

Normals are then transformed into camera space using the current camera transform and alpha composited according to the rendering equation to provide a per-pixel normal estimate:
\begin{equation}
\boldsymbol{\hat {N}} = \sum_{i\in N} {{\boldsymbol{\hat {n}}}}_{i}\alpha_i \prod_{j = 1}^{i - 1} (1- \alpha_j).
\end{equation}

\paragraph{Optimization}
The optimization process is guided by a composite loss function that combines photometric and geometric cues. We adopt the original photometric loss~\cite{kerbl20233d}, which minimizes the difference between the rendered color and the ground truth color of training images. In addition, we regularize the optimization process using the depth maps and surface normals obtained from the geometry enhancement module. The overall loss function is:

\begin{equation}
\mathcal{L} = \mathcal{L}_{\hat{C}} + \lambda_d \mathcal{L}_{\hat{D}} + \lambda_n \mathcal{L}_{\hat{N}},
\label{eq:overall_loss}
\end{equation}
where $\lambda_d$ and $\lambda_n$ are weighting factors to balance the contribution of each term.

To leverage the estimated depth, we minimize a depth loss term:

\begin{equation}
\mathcal{L}_{\hat{D}} = \frac{1}{|\hat{D}|} \sum \log(1 + ||\hat{D} - D||_1),
\label{eq:depth_loss}
\end{equation}
where $\hat{D}$ represents the rendered depth from the 3D Gaussians, and $|\hat{D}|$ indicates the total number of pixels in $\hat{D}$. We found the logarithmic variant of the depth loss yields better results than linear or quadratic penalties, following \cite{turkulainen2025dn}.

We further incorporate a normal consistency term, minimizing the following loss function:

\begin{equation}
\mathcal{L}_{\hat{N}} = \frac{1}{|\hat{N}|} \sum ||\hat{N} - N||_1 + \lambda_s \sum_{i,j} (|\nabla_i \hat{N}_{i,j}| + |\nabla_j \hat{N}_{i,j}|),
\label{eq:normal_loss}
\end{equation}
where $\hat{N}$ represents the rendered surface normal and $\lambda_s$ is a loss weight. The first term promotes alignment between the rendered and estimated normals, while the second enforces smoothness within the rendered normal field.

\paragraph{Compatibility on different 3D reconstruction methods}
By optimizing the 3D Gaussians with both RGB, depth, and normal cues, WonderVerse achieves high-fidelity 3D scene reconstruction that is both photorealistic and geometrically accurate.

In contrast to prior works that rely on point clouds~\cite{yu2024wonderjourney, hollein2023text2room, chung2023luciddreamerdomainfreegeneration3d} or less common data formats such as layered images~\cite{yu2024wonderworld} as intermediate representations, WonderVerse directly generates images and corresponding camera poses. 
This design significantly enhances compatibility, allowing us to leverage well-established 3D reconstruction methods that natively accept image and pose inputs, such as the recent DUSt3R~\cite{dust3r_cvpr24}.  
%
This streamlined approach enables efficient 3D scene generation, with the potential for further improvements as reconstruction techniques advance.

\section{Experiments}
\label{experiments}
\begin{figure*}[h!]
    \centering
        \includegraphics[width=1\linewidth, height=0.25\linewidth]{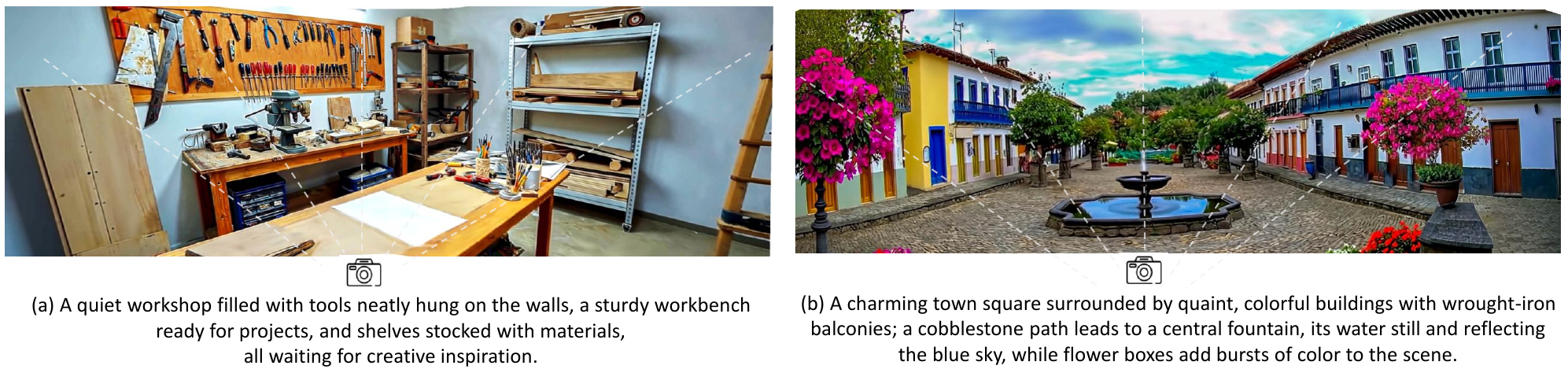}

	\caption{WonderVerse generates large-scale, extendable, coherent, and high-fidelity 3D scenes, both indoors and outdoors.  Dashed lines show the camera’s direction during scene extension. }
	\label{fig:generatedscenes} 
\end{figure*}


\begin{figure*}[h!]
    \centering
    \includegraphics[width=\linewidth , height=0.15\linewidth]{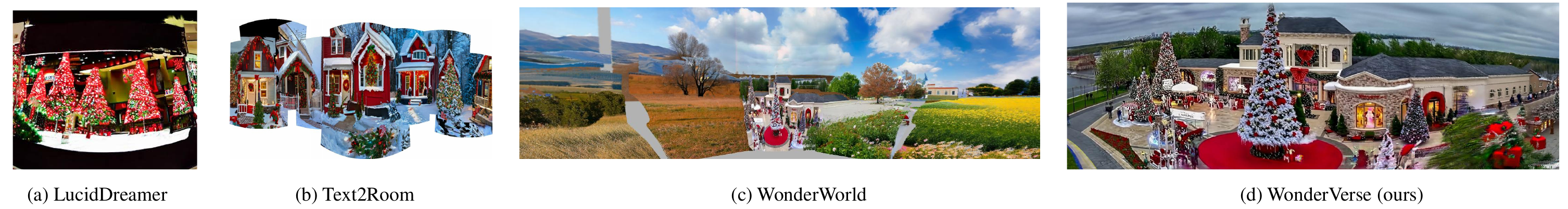}
    \caption{Qualitative comparison with existing works. Text prompt: A small house decorated for Christmas. We can see that LucidDreamer cannot generate a natural scene. Text2Room and WonderWorld suffer from geometric discontinuities, which lead to fragmented scenes and visually inconsistent outputs. In contrast, our WonderVerse can generate high-quality, geometric continuities and natural scenes.}
    \label{fig:comparedgeneratedscenes} 
\end{figure*}

\begin{figure*}[h!]
    \centering
    \includegraphics[width=\linewidth, height=0.35\linewidth]{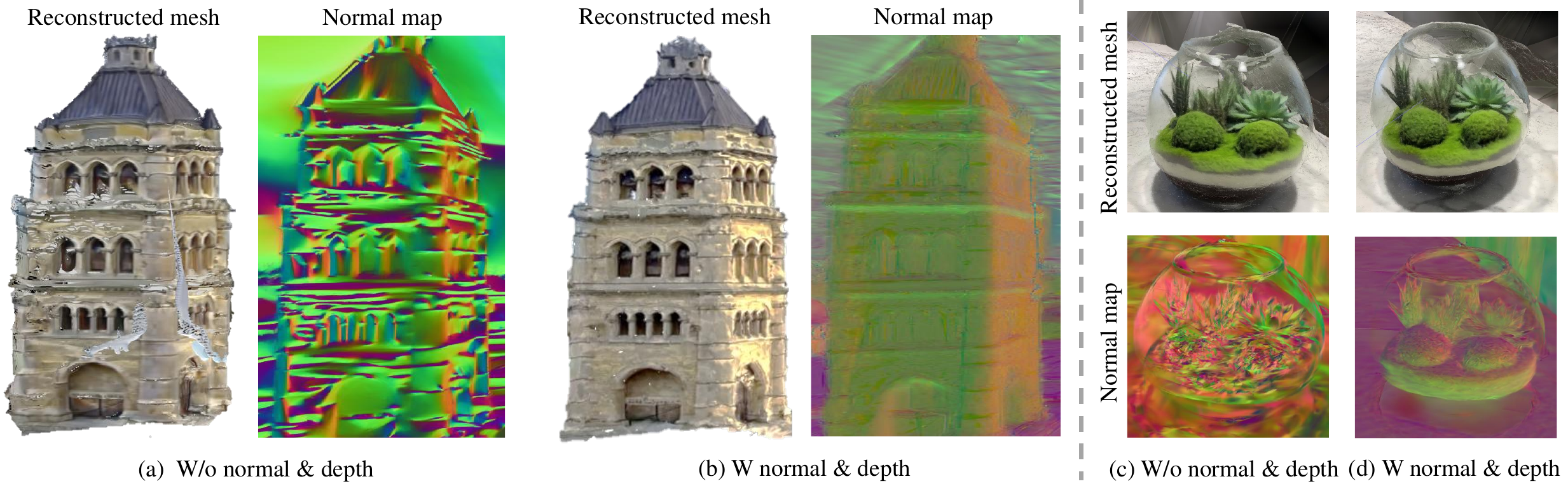}
    \caption{Ablation studies on our geometry enhancement module. 
    }
    \label{fig:Geometry Enhancement}
\end{figure*}

\subsection{Experimental Settings}

\textbf{Implementation Details}
Starting with a text prompt, WonderVerse generates an initial video and then extension videos. Taking the left and right extension as an example, the first and the last frames, along with a text prompt for each, are fed into the video generative model to create two extended videos, repeating this process twice. We set hyper-parameters $T_1$ and $T_2$ as 5 and 0.5 through cross validation. 
Each text prompt starts with the template: “aerial shot, soft lighting, around left [or right], realistic, high-quality, displaying [scene description].”
For initial scene generation, 
we used the Hunyuan video model~\cite{kong2024hunyuanvideo}, and for extensions, we adopted Gen-3 Alpha~\cite{gen3alpha} due to its superior camera control capability. For the LLM-driven extension module, we employed GPT-4o mini~\cite{Achiam2023GPT4TR} to generate camera trajectories conditioned on user-defined scene descriptions. The prompt used is as follows:“Provide the direction and angle of camera pose changes based on the given scene description given by the user. The answer only includes the corresponding angles of up, down, left, right, and the distance of forward and backward movement of four times extend, and it is required that the perspective expansion cannot be repeated. Just give the four answers, format of each answer is: up or down $X$, left or right $X$, forward or back $X$. ($X$ is the angle or distance). If the extension prompt is given, just give one answer. ” 

\noindent\textbf{Evaluation metrics}
Following recent work~\cite{yu2024wonderworld}, we evaluate our scenes using CLIP-Score (CLIP-S)~\cite{hessel2021clipscore}, Q-Align~\cite{wu2023qalign}, and NIQE~\cite{6353522}.  CLIP-S measures text-image alignment by comparing CLIP feature embeddings. Q-Align, trained with human scoring patterns and instruction tuning, assesses the human-aligned generation quality.
NIQE, a reference-free image quality metric, evaluates quality degradation by comparing spatial features to a pristine natural image model.


\subsection{Our Results of Extendable 3D Scenes}
In Figure~\ref{fig:generatedscenes}, we present compelling examples of extendable indoor and outdoor 3D scenes generated by WonderVerse. 
Figure~\ref{fig:generatedscenes} (a, b) illustrate the remarkable realism and rich detail achieved in our created scenes, with flawless geometric integrity.  
Also, the scenes faithfully reflect the text description. 
%
%
%
%
%
This shows the strength of WonderVerse in making highly immersive and extendable 3D scenes.
Besides, we also demonstrate that users can generate extendable 3D scenes iteratively using customized prompts in supplementary material, highlighting the user-controllability of WonderVerse. 

Please also refer to the supplementary material for additional generative results across diverse scenes and styles of WonderVerse. 

\subsection{Comparison with Existing Works}

\textbf{Baseline Approaches}
We compare our WonderVerse with recent works WonderWorld~\cite{yu2024wonderworld} and LucidDreamer~\cite{chung2023luciddreamerdomainfreegeneration3d} both qualitatively and quantitatively. Since these two baselines need an image as input while our method only needs a text prompt as input, for a fair comparison, we take a random frame of our generated video and feed it into their models for 3D scene generation. Note that WonderWorld~\cite{yu2024wonderworld}'s layered image representation, including a sky layer, makes it less suited for indoor scene generation. Consequently, our primary comparative analysis focuses on outdoor scenes.

{Our comparison focuses on \emph{extendable 3D scene generation pipelines} that share the same task setting of \emph{iterative scene expansion} (e.g., LucidDreamer~\cite{chung2023luciddreamerdomainfreegeneration3d} and WonderWorld~\cite{yu2024wonderworld}). Existing methods in this category are predominantly image-driven and rely on iterative image inpainting, making them the most relevant baselines under the same problem formulation.}
{In contrast, video-based approaches (e.g., CameraCtrl~\cite{he2024cameractrl}, SV4D~\cite{xie2024sv4d}, GEN3C~\cite{ren2025gen3c}, GeoVideo~\cite{bai2025geovideo}) mainly target controlled camera motion or bounded video-to-3D reconstruction along a fixed trajectory. WonderVerse addresses a different setting: text-driven and extendable scene generation beyond the original video span; therefore these methods are largely orthogonal to our objective and do not directly support extendable/interactive scene generation.}
 

\textbf{Qualitative Comparison} 
Figure~\ref{fig:comparedgeneratedscenes} visually demonstrates WonderVerse’s clear superiority over existing methods.  As shown in Fig~\ref{fig:comparedgeneratedscenes} (a), LucidDreamer’s generated scene is obscured by odd frame-like artifacts, and lacks detail.  WonderWorld (Figure~\ref{fig:comparedgeneratedscenes} (b, c)) suffers from geometric discontinuities, as indicated by the visibly stitched and unnatural scene. In contrast, WonderVerse (Figure~\ref{fig:comparedgeneratedscenes} (d)) produces a satisfactory and geometrically sound 3D scene. 
The geometric integrity is maintained seamlessly across the scene, addressing the limitations apparent in both LucidDreamer’s quality and WonderWorld’s geometric consistency.
Please refer to the supplementary material for additional comparison results with existing works. 

\textbf{Quantitative Comparison.} We quantitatively compare WonderVerse to prior works LucidDreamer~\cite{chung2023luciddreamerdomainfreegeneration3d} and WonderWorld~\cite{yu2024wonderworld}, with results shown in Table~\ref{tab:qualitative} for both outdoor and indoor scenes. Our method demonstrates state-of-the-art scene generation, outperforming existing methods in semantic alignment, structural consistency, and perceptual quality.  Notably, for outdoor scenes, WonderVerse achieves the best CLIP-S score (0.9219), indicating superior semantic alignment, along with SOTA Q-align and NIQE scores, demonstrating high generation quality. Similarly, for indoor scenes, WonderVerse significantly surpasses LucidDreamer across all metrics, achieving a new SOTA CLIP-S score of 0.9639, again affirming excellent text-scene semantic alignment and high-quality 3D indoor scene generation. Note that WonderWorld cannot generate indoor scenes due to its sky layer, so we do not compare with it in the indoor setting. 

\begin{table}[h!]
\centering
\small
\caption{Evaluation on novel view renderings.}
\label{tab:qualitative}
\scalebox{0.65}{
\begin{tabular}{llcccc}
    \toprule
    Scene Type & Method & Reference  & CLIP-S ($\uparrow$) & Q-align ($\uparrow$) & NIQE ($\uparrow$) \\
    \midrule
    \multirow{3}{*}{Outdoor} 
    & 
    LucidDreamer & arXiv24 & 0.4200 & 2.6704 & 0.7786 \\
    & WonderWorld & CVPR25 & 0.7817 & 2.8302 & 0.7127 \\ 
    & WonderVerse (ours) & - & \textbf{0.9219} & \textbf{3.2229} & \textbf{0.7806} \\
    \midrule
    \multirow{2}{*}{Indoor} 
    & LucidDreamer & arXiv24 & 0.4633 & 3.9870 & 0.6655 \\ 
    & WonderVerse (ours) & - & \textbf{0.9639} & \textbf{4.2680} & \textbf{0.7119} \\
    \bottomrule
  \end{tabular}
}
\end{table}

\begin{figure}[h!]
    \centering
    \subfloat[]{%
        \includegraphics[width=0.45\linewidth]{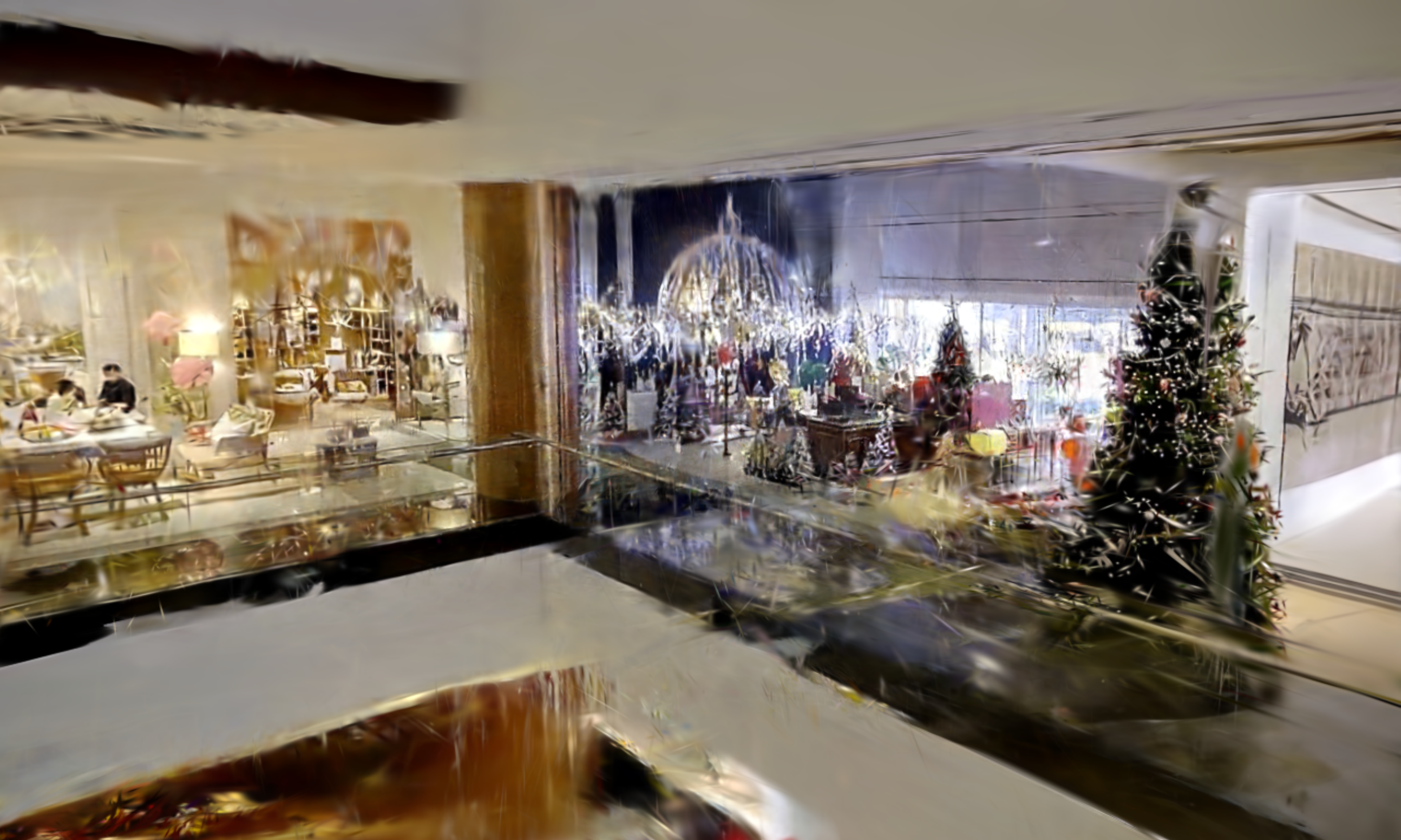}
    }
    \hspace{3mm}
    \subfloat[]{%
        \includegraphics[width=0.45\linewidth]{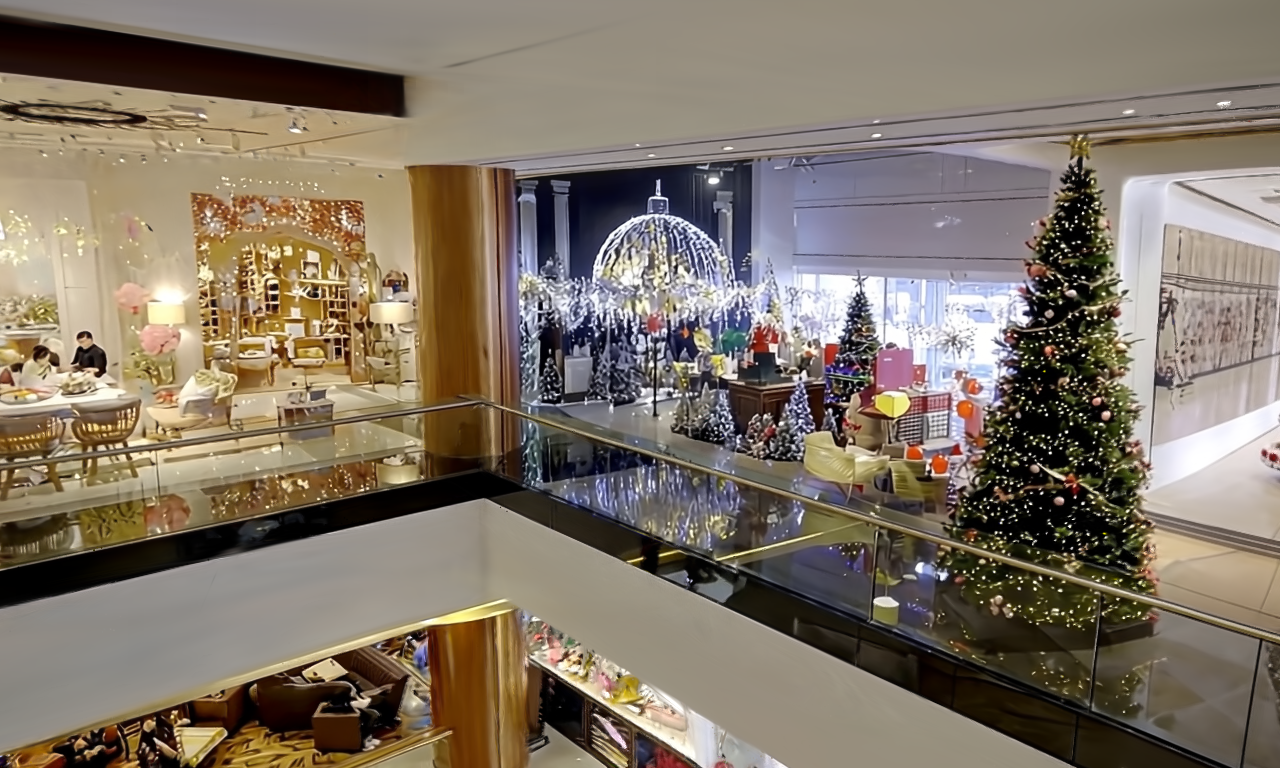}
    }
    \caption{Ablation study on our abnormal sequence detection module. (a) W/o abnormal sequence detection; (b) W abnormal sequence detection
    }
    \label{fig:abnormal_detection}
\end{figure}

\subsection{Ablation Studies}
\label{Ablations}
In this section, we examine our key modules, including abnormal sequence detection and geometric enhancement. In the supplementary material, we also examine scene extension module and compare different 3D representation methods 3DGS~\cite{kerbl20233d} and DUSt3R~\cite{dust3r_cvpr24} to demonstrate the compatibility of our approach.

\subsubsection{Abnormal Sequence Detection}
In this section, we study the effectiveness of our abnormal sequence detection. As illustrated in Figure~\ref{fig:abnormal_detection}, 
without our detection algorithm, noticeable artifacts degrade the generated 3D scene due to the geometric inconsistency of the generated video. In contrast, when abnormal sequences are detected and addressed, the generative quality is much higher. 
The improvement is further demonstrated by quantitative evaluations, please refer to the supplementary material for more details. 


\subsubsection{Geometric Enhancement}
As illustrated in Figure~\ref{fig:Geometry Enhancement}, we present 3D mesh reconstructions with and without the geometry enhancement module. Without depth and normal regularization, the tower exhibits considerable surface distortion, and the bowl displays noticeable artifacts. Conversely, incorporating geometric priors significantly improves surface quality; for instance, the tower’s intricate details are faithfully generated, and the glass bowl’s surface is much smoother. The normal maps further demonstrate that the geometry enhancement module enables a cleaner 3D surface and creates complex geometric details.

\section{Limitations and Future Work}
\label{sec:limitations}
Our WonderVerse framework, while effective, has certain limitations. First, The quality of our simulated 3D environments is inherently limited by existing video generation models.
{Second, our abnormal sequence detection is a heuristic filtering mechanism based on SfM (COLMAP) pose smoothness. It can miss subtle geometric warping or trigger false positives under challenging conditions (e.g., repetitive textures or weak features).}
Finally, our model mainly focuses on static 3D scene generation due to the use of 3DGS and DUSt3R for reconstruction. To overcome this, an exciting avenue for future research is to combine with recent advancements in dynamic scene representation \cite{Duan20244DRotorGS,Yang2023RealtimePD,Wu20234DGS} to pave the way for generating dynamic and animated 3D environments.
We also provide representative failure cases and analysis in the supplementary material.

\section{Conclusion}
\label{Conclusion and Limitation}

In this paper, we introduce WonderVerse, a simple yet effective framework for 3D scene generation.
Unlike previous methods that rely on image-based generation schemes, we leverage the world-level priors embedded within video generative models to improve the generation quality. 
{Our contribution lies in demonstrating how recent video generative priors can be reliably harnessed for extendable 3D scene generation, a setting where existing methods struggle. Our work bridges video generation and usable 3D scene reconstruction, enabling extenable scene generation. In particular, prior extendable scene generation pipelines are predominantly image-based and often suffer from geometric distortion, error accumulation, and seam artifacts during iterative expansion.}
Equipped with a video geometric correction module that detects abnormal videos due to extreme camera pose changes and a 3D geometric enhancement module that ensures depth and normal consistency, WonderVerse can generate extendable and highly realistic 3D scenes, markedly outperforming existing works with more complex architectures.
%
%
Please refer to the supplementary material for discussions of limitations.

\section*{Acknowledgement}

This work has been supported by National Natural Science Foundation of China (Category C) fund code 62506149 , Lingnan University StartUp Grant fund code: SUG-001/2526, and Faculty Research Grant fund code:106106 and 106119.


{\small
\bibliographystyle{cvm}
\bibliography{cvmbib}

@article{liu2023exim,
  title={EXIM: A Hybrid Explicit-Implicit Representation for Text-Guided 3D Shape Generation},
  author={Liu, Zhengzhe and Hu, Jingyu and Hui, Ka-Hei and Qi, Xiaojuan and Cohen-Or, Daniel and Fu, Chi-Wing},
  journal={ACM Trans. Graph.},
  volume={42},
  number={6},
  pages={1--12},
  year={2023}
}

@article{yan2026comp,
  title={LaS-Comp: Zero-shot 3D Completion with Latent-Spatial Consistency},
  author={Yan, Weilong and Li, Haipeng and Xu, Hao and Ye, Nianjin and Ai, Yihao and Liu, Shuaicheng and Hu, Jingyu},
  journal={arXiv preprint arXiv:2602.18735},
  year={2026}
}

@article{hu2023neural,
  title={Neural Wavelet-domain Diffusion for 3D Shape Generation, Inversion, and Manipulation},
  author={Hu, Jingyu and Hui, Ka-Hei and Liu, Zhengzhe and Li, Ruihui and Fu, Chi-Wing},
  journal={ACM Transactions on Graphics (TOG)},
  year    = {2024},
  volume  = {42},
  number  = {6}
}

@article{hu2026pegasus3dpersonalizationgeometry,
      title={PEGAsus: 3D Personalization of Geometry and Appearance}, 
      author={Jingyu Hu and Bin Hu and Ka-Hei Hui and Haipeng Li and Zhengzhe Liu and Daniel Cohen-Or and Chi-Wing Fu},
      year={2026},
      journal={arXiv preprint arXiv:2602.08198},
}

@inproceedings{hui2022template,
                title = {Neural Template: Topology-aware Reconstruction and Disentangled Generation of 3D Meshes},
                author = {Ka-Hei Hui and Ruihui Li and Jingyu Hu and Chi-Wing Fu},
                booktitle={Proceedings of the IEEE/CVF conference on computer vision and pattern recognition},
                 pages = {18572-18582},
                year={2022}
                }

@inproceedings{hui2022neural,
  title={Neural Wavelet-domain Diffusion for 3D Shape Generation},
  author={Hui, Ka-Hei and Li, Ruihui and Hu, Jingyu and Fu, Chi-Wing},
  booktitle={Proceedings of SIGGRAPH Asia},
  pages={1--9},
  year={2022}
}

@inproceedings{du2025hierarchical,
  title={Hierarchical neural semantic representation for 3d semantic correspondence},
  author={Du, Keyu and Hu, Jingyu and Li, Haipeng and Xu, Hao and Huang, Haibin and Fu, Chi-Wing and Liu, Shuaicheng},
  booktitle={Proceedings of SIGGRAPH Asia},
  pages={1--11},
  year={2025}
}

@inproceedings{hu2024_cnsedit,
  title     = {CNS-Edit: 3D Shape Editing via Coupled Neural Shape Optimization},
  author    = {Hu, Jingyu and Hui, Ka-Hei and Liu, Zhengzhe and Zhang, Hao and Fu, Chi-Wing},
  booktitle = {Proceedings of SIGGRAPH},
  year      = {2024},
  pages     = {1--12}
}

@inproceedings{hu2023clipxplore,
  author = {Hu, Jingyu and Hui, Ka-Hei and Liu, Zhengzhe and Zhang, Hao and Fu, Chi-Wing},
  title = {CLIPXPlore: Coupled CLIP and Shape Spaces for 3D Shape Exploration},
  year = {2023},
  booktitle = {Proceedings of SIGGRAPH Asia},
  pages={1--12}
}

@article{kerbl20233d,
  title={3d gaussian splatting for real-time radiance field rendering.},
  author={Kerbl, Bernhard and Kopanas, Georgios and Leimk{\"u}hler, Thomas and Drettakis, George},
  journal={ACM Trans. Graph.},
  volume={42},
  number={4},
  pages={139--1},
  year={2023}
}

@misc{chung2023luciddreamerdomainfreegeneration3d,
      title={LucidDreamer: Domain-free Generation of 3D Gaussian Splatting Scenes}, 
      author={Jaeyoung Chung and Suyoung Lee and Hyeongjin Nam and Jaerin Lee and Kyoung Mu Lee},
      year={2023},
      eprint={2311.13384},
      archivePrefix={arXiv},
      primaryClass={cs.CV},
      url={https://arxiv.org/abs/2311.13384}, 
}

@inproceedings{hollein2023text2room,
  title={Text2room: Extracting textured 3d meshes from 2d text-to-image models},
  author={H{\"o}llein, Lukas and Cao, Ang and Owens, Andrew and Johnson, Justin and Nie{\ss}ner, Matthias},
  booktitle={Proceedings of the IEEE/CVF International Conference on Computer Vision},
  pages={7909--7920},
  year={2023}
}

@article{yu2024wonderworld,
  title={WonderWorld: Interactive 3D Scene Generation from a Single Image},
  author={Yu, Hong-Xing and Duan, Haoyi and Herrmann, Charles and Freeman, William T and Wu, Jiajun},
  journal={arXiv preprint arXiv:2406.09394},
  year={2024}
}

@article{yang2024depth_v2,
  title={Depth anything v2},
  author={Yang, Lihe and Kang, Bingyi and Huang, Zilong and Zhao, Zhen and Xu, Xiaogang and Feng, Jiashi and Zhao, Hengshuang},
  journal={Advances in Neural Information Processing Systems},
  volume={37},
  pages={21875--21911},
  year={2024}
}

@inproceedings{yin2023metric3d,
  title={Metric3d: Towards zero-shot metric 3d prediction from a single image},
  author={Yin, Wei and Zhang, Chi and Chen, Hao and Cai, Zhipeng and Yu, Gang and Wang, Kaixuan and Chen, Xiaozhi and Shen, Chunhua},
  booktitle={Proceedings of the IEEE/CVF International Conference on Computer Vision},
  pages={9043--9053},
  year={2023}
}

@inproceedings{chen2025video,
  title={Video Depth Anything: Consistent Depth Estimation for Super-Long Videos},
  author={Chen, Sili and Guo, Hengkai and Zhu, Shengnan and Zhang, Feihu and Huang, Zilong and Feng, Jiashi and Kang, Bingyi},
  booktitle={Proceedings of the IEEE/CVF International Conference on Computer Vision},
  year={2025}
}

@inproceedings{hu2024depthcrafter,
  title={Depthcrafter: Generating consistent long depth sequences for open-world videos},
  author={Hu, Wenbo and Gao, Xiangjun and Li, Xiaoyu and Zhao, Sijie and Cun, Xiaodong and Zhang, Yong and Quan, Long and Shan, Ying},
  booktitle={Proceedings of the IEEE/CVF International Conference on Computer Vision},
  year={2025}
}

@article{ye2024stablenormal,
  title={Stablenormal: Reducing diffusion variance for stable and sharp normal},
  author={Ye, Chongjie and Qiu, Lingteng and Gu, Xiaodong and Zuo, Qi and Wu, Yushuang and Dong, Zilong and Bo, Liefeng and Xiu, Yuliang and Han, Xiaoguang},
  journal={ACM Transactions on Graphics (TOG)},
  volume={43},
  number={6},
  pages={1--18},
  year={2024}
}

@inproceedings{qiu2024richdreamer,
  title={Richdreamer: A generalizable normal-depth diffusion model for detail richness in text-to-3d},
  author={Qiu, Lingteng and Chen, Guanying and Gu, Xiaodong and Zuo, Qi and Xu, Mutian and Wu, Yushuang and Yuan, Weihao and Dong, Zilong and Bo, Liefeng and Han, Xiaoguang},
  booktitle={Proceedings of the IEEE/CVF conference on computer vision and pattern recognition},
  pages={9914--9925},
  year={2024}
}

@inproceedings{bae2024rethinking,
  title={Rethinking inductive biases for surface normal estimation},
  author={Bae, Gwangbin and Davison, Andrew J},
  booktitle={Proceedings of the IEEE/CVF Conference on Computer Vision and Pattern Recognition},
  pages={9535--9545},
  year={2024}
}

@article{hu2024metric3d,
  title={Metric3d v2: A versatile monocular geometric foundation model for zero-shot metric depth and surface normal estimation},
  author={Hu, Mu and Yin, Wei and Zhang, Chi and Cai, Zhipeng and Long, Xiaoxiao and Chen, Hao and Wang, Kaixuan and Yu, Gang and Shen, Chunhua and Shen, Shaojie},
  journal={IEEE Transactions on Pattern Analysis and Machine Intelligence},
  year={2024},
  publisher={IEEE}
}

@inproceedings{fu2024geowizard,
  title={Geowizard: Unleashing the diffusion priors for 3d geometry estimation from a single image},
  author={Fu, Xiao and Yin, Wei and Hu, Mu and Wang, Kaixuan and Ma, Yuexin and Tan, Ping and Shen, Shaojie and Lin, Dahua and Long, Xiaoxiao},
  booktitle={European Conference on Computer Vision},
  pages={241--258},
  year={2024},
  organization={Springer}
}

@inproceedings{ke2024repurposing,
  title={Repurposing diffusion-based image generators for monocular depth estimation},
  author={Ke, Bingxin and Obukhov, Anton and Huang, Shengyu and Metzger, Nando and Daudt, Rodrigo Caye and Schindler, Konrad},
  booktitle={Proceedings of the IEEE/CVF Conference on Computer Vision and Pattern Recognition},
  pages={9492--9502},
  year={2024}
}

@inproceedings{liu2023zero,
  title={Zero-1-to-3: Zero-shot one image to 3d object},
  author={Liu, Ruoshi and Wu, Rundi and Van Hoorick, Basile and Tokmakov, Pavel and Zakharov, Sergey and Vondrick, Carl},
  booktitle={Proceedings of the IEEE/CVF international conference on computer vision},
  pages={9298--9309},
  year={2023}
}

@misc{kong2024hunyuanvideo,
      title={HunyuanVideo: A Systematic Framework For Large Video Generative Models}, 
      author={Weijie Kong},
      year={2024},
      archivePrefix={arXiv preprint arXiv:2412.03603},
      primaryClass={cs.CV},
      url={https://arxiv.org/abs/2412.03603}, 
}

@inproceedings{ye2025hi3dgen,
  title={Hi3DGen: High-fidelity 3D Geometry Generation from Images via Normal Bridging},
  author={Ye, Chongjie and Wu, Yushuang and Lu, Ziteng and Chang, Jiahao and Guo, Xiaoyang and Zhou, Jiaqing and Zhao, Hao and Han, Xiaoguang},
  booktitle={Conference on Computer Vision and Pattern Recognition (CVPR)},
  year={2025}
}

@inproceedings{dust3r_cvpr24,
      title={DUSt3R: Geometric 3D Vision Made Easy}, 
      author={Shuzhe Wang and Vincent Leroy and Yohann Cabon and Boris Chidlovskii and Jerome Revaud},
      booktitle = {CVPR},
      year = {2024}
}

@inproceedings{bar2024lumiere,
  title={Lumiere: A space-time diffusion model for video generation},
  author={Bar-Tal, Omer and Chefer, Hila and Tov, Omer and Herrmann, Charles and Paiss, Roni and Zada, Shiran and Ephrat, Ariel and Hur, Junhwa and Liu, Guanghui and Raj, Amit and others},
  booktitle={SIGGRAPH Asia 2024 Conference Papers},
  pages={1--11},
  year={2024}
}

@article{vondrick2016generating,
  title={Generating videos with scene dynamics},
  author={Vondrick, Carl and Pirsiavash, Hamed and Torralba, Antonio},
  journal={Advances in neural information processing systems},
  volume={29},
  year={2016}
}

@inproceedings{li2024dreamscene,
  title={Dreamscene: 3d gaussian-based text-to-3d scene generation via formation pattern sampling},
  author={Li, Haoran and Shi, Haolin and Zhang, Wenli and Wu, Wenjun and Liao, Yong and Wang, Lin and Lee, Lik-hang and Zhou, Peng Yuan},
  booktitle={European Conference on Computer Vision},
  pages={214--230},
  year={2024},
  organization={Springer}
}

@inproceedings{rombach2022high,
  title={High-resolution image synthesis with latent diffusion models},
  author={Rombach, Robin and Blattmann, Andreas and Lorenz, Dominik and Esser, Patrick and Ommer, Bj{\"o}rn},
  booktitle={Proceedings of the IEEE/CVF conference on computer vision and pattern recognition},
  pages={10684--10695},
  year={2022}
}

@inproceedings{voleti2024sv3d,
  title={Sv3d: Novel multi-view synthesis and 3d generation from a single image using latent video diffusion},
  author={Voleti, Vikram and Yao, Chun-Han and Boss, Mark and Letts, Adam and Pankratz, David and Tochilkin, Dmitry and Laforte, Christian and Rombach, Robin and Jampani, Varun},
  booktitle={European Conference on Computer Vision},
  pages={439--457},
  year={2024},
  organization={Springer}
}

@article{xie2024sv4d,
  title={Sv4d: Dynamic 3d content generation with multi-frame and multi-view consistency},
  author={Xie, Yiming and Yao, Chun-Han and Voleti, Vikram and Jiang, Huaizu and Jampani, Varun},
  journal={arXiv preprint arXiv:2407.17470},
  year={2024}
}

@inproceedings{hessel2021clipscore,
  title={{CLIPScore:} A Reference-free Evaluation Metric for Image Captioning},
  author={Hessel, Jack and Holtzman, Ari and Forbes, Maxwell and Bras, Ronan Le and Choi, Yejin},
  booktitle={EMNLP},
  year={2021}
}

@article{wu2023qalign,
  title={Q-Align: Teaching LMMs for Visual Scoring via Discrete Text-Defined Levels},
  author={Wu, Haoning and Zhang, Zicheng and Zhang, Weixia and Chen, Chaofeng and Li, Chunyi and Liao, Liang and Wang, Annan and Zhang, Erli and Sun, Wenxiu and Yan, Qiong and Min, Xiongkuo and Zhai, Guangtai and Lin, Weisi},
  journal={arXiv preprint arXiv:2312.17090},
  year={2023},
  institution={Nanyang Technological University and Shanghai Jiao Tong University and Sensetime Research},
  note={Equal Contribution by Wu, Haoning and Zhang, Zicheng. Project Lead by Wu, Haoning. Corresponding Authors: Zhai, Guangtai and Lin, Weisi.}
}

@ARTICLE{6353522,
  author={Mittal, Anish and Soundararajan, Rajiv and Bovik, Alan C.},
  journal={IEEE Signal Processing Letters}, 
  title={Making a “Completely Blind” Image Quality Analyzer}, 
  year={2013},
  volume={20},
  number={3},
  pages={209-212},
  doi={10.1109/LSP.2012.2227726}}

@inproceedings{yu2024wonderjourney,
  title={Wonderjourney: Going from anywhere to everywhere},
  author={Yu, Hong-Xing and Duan, Haoyi and Hur, Junhwa and Sargent, Kyle and Rubinstein, Michael and Freeman, William T and Cole, Forrester and Sun, Deqing and Snavely, Noah and Wu, Jiajun and others},
  booktitle={Proceedings of the IEEE/CVF Conference on Computer Vision and Pattern Recognition},
  pages={6658--6667},
  year={2024}
}

@misc{zeng2023makepixelsdancehighdynamic,
      title={Make Pixels Dance: High-Dynamic Video Generation}, 
      author={Yan Zeng and Guoqiang Wei and Jiani Zheng and Jiaxin Zou and Yang Wei and Yuchen Zhang and Hang Li},
      year={2023},
      eprint={2311.10982},
      archivePrefix={arXiv},
      primaryClass={cs.CV},
      url={https://arxiv.org/abs/2311.10982}, 
}

@article{chen2025goku, title={Goku: Flow Based Video Generative Foundation Models}, author={Chen, Shoufa and Ge, Chongjian and Zhang, Yuqi and Zhang, Yida and Zhu, Fengda and Yang, Hao and Hao, Hongxiang and Wu, Hui and Lai, Zhichao and Hu, Yifei and Lin, Ting-Che and Zhang, Shilong and Li, Fu and Li, Chuan and Wang, Xing and Peng, Yanghua and Sun, Peize and Luo, Ping and Jiang, Yi and Yuan, Zehuan and Peng, Bingyue and Liu, Xiaobing}, journal={arXiv preprint arXiv:2502.04896}, year={2025} }

@article{zheng2024open,
  title={Open-sora: Democratizing efficient video production for all},
  author={Zheng, Zangwei and Peng, Xiangyu and Yang, Tianji and Shen, Chenhui and Li, Shenggui and Liu, Hongxin and Zhou, Yukun and Li, Tianyi and You, Yang},
  journal={arXiv preprint arXiv:2412.20404},
  year={2024}
}

@article{yang2024cogvideox,
  title={Cogvideox: Text-to-video diffusion models with an expert transformer},
  author={Yang, Zhuoyi and Teng, Jiayan and Zheng, Wendi and Ding, Ming and Huang, Shiyu and Xu, Jiazheng and Yang, Yuanming and Hong, Wenyi and Zhang, Xiaohan and Feng, Guanyu and others},
  journal={arXiv preprint arXiv:2408.06072},
  year={2024}
}

@article{qin2024worldsimbench,
  title={Worldsimbench: Towards video generation models as world simulators},
  author={Qin, Yiran and Shi, Zhelun and Yu, Jiwen and Wang, Xijun and Zhou, Enshen and Li, Lijun and Yin, Zhenfei and Liu, Xihui and Sheng, Lu and Shao, Jing and others},
  journal={arXiv preprint arXiv:2410.18072},
  year={2024}
}

@article{chang2024matters,
  title={What Matters in Detecting AI-Generated Videos like Sora?},
  author={Chang, Chirui and Liu, Zhengzhe and Lyu, Xiaoyang and Qi, Xiaojuan},
  journal={arXiv preprint arXiv:2406.19568},
  year={2024}
}

@article{he2024cameractrl,
  title={Cameractrl: Enabling camera control for text-to-video generation},
  author={He, Hao and Xu, Yinghao and Guo, Yuwei and Wetzstein, Gordon and Dai, Bo and Li, Hongsheng and Yang, Ceyuan},
  journal={arXiv preprint arXiv:2404.02101},
  year={2024}
}

@inproceedings{wang2024motionctrl,
  title={Motionctrl: A unified and flexible motion controller for video generation},
  author={Wang, Zhouxia and Yuan, Ziyang and Wang, Xintao and Li, Yaowei and Chen, Tianshui and Xia, Menghan and Luo, Ping and Shan, Ying},
  booktitle={ACM SIGGRAPH 2024 Conference Papers},
  pages={1--11},
  year={2024}
}

@inproceedings{yang2024depth,
  title={Depth anything: Unleashing the power of large-scale unlabeled data},
  author={Yang, Lihe and Kang, Bingyi and Huang, Zilong and Xu, Xiaogang and Feng, Jiashi and Zhao, Hengshuang},
  booktitle={Proceedings of the IEEE/CVF Conference on Computer Vision and Pattern Recognition},
  pages={10371--10381},
  year={2024}
}

@inproceedings{long2024wonder3d,
  title={Wonder3d: Single image to 3d using cross-domain diffusion},
  author={Long, Xiaoxiao and Guo, Yuan-Chen and Lin, Cheng and Liu, Yuan and Dou, Zhiyang and Liu, Lingjie and Ma, Yuexin and Zhang, Song-Hai and Habermann, Marc and Theobalt, Christian and others},
  booktitle=CVPR,
  pages={9970--9980},
  year={2024}
}

@article{gao2024cat3d,
  title={Cat3d: Create anything in 3d with multi-view diffusion models},
  author={Gao, Ruiqi and Holynski, Aleksander and Henzler, Philipp and Brussee, Arthur and Martin-Brualla, Ricardo and Srinivasan, Pratul and Barron, Jonathan T and Poole, Ben},
  journal={arXiv preprint arXiv:2405.10314},
  year={2024}
}

@inproceedings{wu2024reconfusion,
  title={Reconfusion: 3d reconstruction with diffusion priors},
  author={Wu, Rundi and Mildenhall, Ben and Henzler, Philipp and Park, Keunhong and Gao, Ruiqi and Watson, Daniel and Srinivasan, Pratul P and Verbin, Dor and Barron, Jonathan T and Poole, Ben and others},
  booktitle={Proceedings of the IEEE/CVF conference on computer vision and pattern recognition},
  pages={21551--21561},
  year={2024}
}

@inproceedings{lin2023infinicity,
  title={Infinicity: Infinite-scale city synthesis},
  author={Lin, Chieh Hubert and Lee, Hsin-Ying and Menapace, Willi and Chai, Menglei and Siarohin, Aliaksandr and Yang, Ming-Hsuan and Tulyakov, Sergey},
  booktitle={Proceedings of the IEEE/CVF international conference on computer vision},
  pages={22808--22818},
  year={2023}
}

@inproceedings{xie2024citydreamer,
  title={Citydreamer: Compositional generative model of unbounded 3d cities},
  author={Xie, Haozhe and Chen, Zhaoxi and Hong, Fangzhou and Liu, Ziwei},
  booktitle={Proceedings of the IEEE/CVF conference on computer vision and pattern recognition},
  pages={9666--9675},
  year={2024}
}

@article{deng2023citygen,
  title={Citygen: Infinite and controllable 3d city layout generation},
  author={Deng, Jie and Chai, Wenhao and Guo, Jianshu and Huang, Qixuan and Hu, Wenhao and Hwang, Jenq-Neng and Wang, Gaoang},
  journal={arXiv preprint arXiv:2312.01508},
  year={2023}
}

@article{wu2024blockfusion,
  title={Blockfusion: Expandable 3d scene generation using latent tri-plane extrapolation},
  author={Wu, Zhennan and Li, Yang and Yan, Han and Shang, Taizhang and Sun, Weixuan and Wang, Senbo and Cui, Ruikai and Liu, Weizhe and Sato, Hiroyuki and Li, Hongdong and others},
  journal={ACM Transactions on Graphics (TOG)},
  volume={43},
  number={4},
  pages={1--17},
  year={2024},
  publisher={ACM New York, NY, USA}
}

@article{chen2023scenedreamer,
  title={Scenedreamer: Unbounded 3d scene generation from 2d image collections},
  author={Chen, Zhaoxi and Wang, Guangcong and Liu, Ziwei},
  journal={IEEE transactions on pattern analysis and machine intelligence},
  volume={45},
  number={12},
  pages={15562--15576},
  year={2023},
  publisher={IEEE}
}

@inproceedings{chai2023persistent,
  title={Persistent nature: A generative model of unbounded 3d worlds},
  author={Chai, Lucy and Tucker, Richard and Li, Zhengqi and Isola, Phillip and Snavely, Noah},
  booktitle={Proceedings of the IEEE/CVF conference on computer vision and pattern recognition},
  pages={20863--20874},
  year={2023}
}

@article{xie2024gaussiancity,
  title={GaussianCity: Generative Gaussian splatting for unbounded 3D city generation},
  author={Xie, Haozhe and Chen, Zhaoxi and Hong, Fangzhou and Liu, Ziwei},
  journal={arXiv preprint arXiv:2406.06526},
  year={2024}
}

@article{zhang2024cityx,
  title={Cityx: Controllable procedural content generation for unbounded 3d cities},
  author={Zhang, Shougao and Zhou, Mengqi and Wang, Yuxi and Luo, Chuanchen and Wang, Rongyu and Li, Yiwei and Zhang, Zhaoxiang and Peng, Junran},
  journal={arXiv preprint arXiv:2407.17572},
  year={2024}
}

@inproceedings{liu2024pyramid,
  title={Pyramid diffusion for fine 3d large scene generation},
  author={Liu, Yuheng and Li, Xinke and Li, Xueting and Qi, Lu and Li, Chongshou and Yang, Ming-Hsuan},
  booktitle={European Conference on Computer Vision},
  pages={71--87},
  year={2024},
  organization={Springer}
}

@article{wei2024planner3d,
  title={Planner3D: LLM-enhanced graph prior meets 3D indoor scene explicit regularization},
  author={Wei, Yao and Min, Martin Renqiang and Vosselman, George and Li, Li Erran and Yang, Michael Ying},
  journal={arXiv preprint arXiv:2403.12848},
  year={2024}
}

@article{fridman2023scenescape,
  title={Scenescape: Text-driven consistent scene generation},
  author={Fridman, Rafail and Abecasis, Amit and Kasten, Yoni and Dekel, Tali},
  journal={Advances in Neural Information Processing Systems},
  volume={36},
  pages={39897--39914},
  year={2023}
}

@article{gen3alpha,
  title={https://runwayml.com/research/introducing-gen-3-alpha},
  author={Runway},
  year={2024}
}

@inproceedings{Duan20244DRotorGS,
  title={4D-Rotor Gaussian Splatting: Towards Efficient Novel View Synthesis for Dynamic Scenes},
  author={Yuanxing Duan and Fangyin Wei and Qiyu Dai and Yuhang He and Wenzheng Chen and Baoquan Chen},
  booktitle={International Conference on Computer Graphics and Interactive Techniques},
  year={2024},
  url={https://api.semanticscholar.org/CorpusID:267411895}
}

@inproceedings{Yang2023RealtimePD,
  title={Real-time Photorealistic Dynamic Scene Representation and Rendering with 4D Gaussian Splatting},
  author={Yang, Zeyu and Yang, Hongye and Pan, Zijie and Zhang, Li},
  booktitle = {International Conference on Learning Representations (ICLR)},
  year={2024}
}

@article{Wu20234DGS,
  title={4D Gaussian Splatting for Real-Time Dynamic Scene Rendering},
  author={Guanjun Wu and Taoran Yi and Jiemin Fang and Lingxi Xie and Xiaopeng Zhang and Wei Wei and Wenyu Liu and Qi Tian and Xinggang Wang},
  journal={2024 IEEE/CVF Conference on Computer Vision and Pattern Recognition (CVPR)},
  year={2023},
  pages={20310-20320},
  url={https://api.semanticscholar.org/CorpusID:263908793}
}

@inproceedings{ranftl2021vision,
  title={Vision transformers for dense prediction},
  author={Ranftl, Ren{\'e} and Bochkovskiy, Alexey and Koltun, Vladlen},
  booktitle={Proceedings of the IEEE/CVF international conference on computer vision},
  pages={12179--12188},
  year={2021}
}

@inproceedings{turkulainen2025dn,
  title={Dn-splatter: Depth and normal priors for gaussian splatting and meshing},
  author={Turkulainen, Matias and Ren, Xuqian and Melekhov, Iaroslav and Seiskari, Otto and Rahtu, Esa and Kannala, Juho},
  booktitle={2025 IEEE/CVF Winter Conference on Applications of Computer Vision (WACV)},
  pages={2421--2431},
  year={2025},
  organization={IEEE}
}

@article{lee2024vividdream,
  title={Vividdream: Generating 3d scene with ambient dynamics},
  author={Lee, Yao-Chih and Chen, Yi-Ting and Wang, Andrew and Liao, Ting-Hsuan and Feng, Brandon Y and Huang, Jia-Bin},
  journal={arXiv preprint arXiv:2405.20334},
  year={2024}
}

@article{blattmann2023stable,
  title={Stable video diffusion: Scaling latent video diffusion models to large datasets},
  author={Blattmann, Andreas and Dockhorn, Tim and Kulal, Sumith and Mendelevitch, Daniel and Kilian, Maciej and Lorenz, Dominik and Levi, Yam and English, Zion and Voleti, Vikram and Letts, Adam and others},
  journal={arXiv preprint arXiv:2311.15127},
  year={2023}
}

@inproceedings{Achiam2023GPT4TR,
  title={GPT-4 Technical Report},
  author={OpenAI Josh Achiam and Steven Adler and Sandhini Agarwal},
  year={2023},
  url={https://api.semanticscholar.org/CorpusID:257532815}
}

@inproceedings{ren2025gen3c,
  title={Gen3c: 3d-informed world-consistent video generation with precise camera control},
  author={Ren, Xuanchi and Shen, Tianchang and Huang, Jiahui and Ling, Huan and Lu, Yifan and Nimier-David, Merlin and M{\"u}ller, Thomas and Keller, Alexander and Fidler, Sanja and Gao, Jun},
  booktitle={Proceedings of the Computer Vision and Pattern Recognition Conference},
  pages={6121--6132},
  year={2025}
}

@article{bai2025geovideo,
  title={Geovideo: Introducing geometric regularization into video generation model},
  author={Bai, Yunpeng and Fang, Shaoheng and Yu, Chaohui and Wang, Fan and Huang, Qixing},
  journal={arXiv preprint arXiv:2512.03453},
  year={2025}
}
}

\end{document}